\documentclass[a4paper, 10pt, conference]{ieeeconf}      

\IEEEoverridecommandlockouts                              

\usepackage{graphics} 
\usepackage{epsfig} 
\usepackage{mathptmx} 
\usepackage{times} 
\usepackage{amsmath} 
\usepackage{amssymb}  

\newcommand{\norm}[1]{\lVert#1\rVert}

\title{\LARGE \bf
 Towards Unsupervised Learning for Instrument Segmentation in Robotic Surgery with Cycle-Consistent Adversarial Networks
}

\author{Daniil Pakhomov$^{1}$ Wei Shen$^{1}$  and Nassir Navab$^{1}$
\thanks{$^{1}$Johns Hopkins University, Baltimore, USA}
}

\begin{document}

\maketitle
\thispagestyle{empty}
\pagestyle{empty}

\begin{abstract}

Surgical tool segmentation in endoscopic images is an important problem: it is a crucial step towards full instrument pose estimation and it is used for integration of pre- and intra-operative images into the endoscopic view. While many recent approaches based on convolutional
neural networks have shown great results, a key barrier
to progress lies in the acquisition of a
large number of manually-annotated images which is necessary for an algorithm to generalize and work well in diverse surgical scenarios. Unlike the surgical image data itself, annotations are difficult to acquire and may be of variable quality. On the other hand, synthetic annotations can be automatically generated by using forward kinematic model of the robot and CAD models of tools by projecting them onto an image plane. Unfortunately, this model is very inaccurate and cannot be used for supervised learning of image segmentation models.  Since generated annotations will not directly correspond to endoscopic images due to errors, we formulate the problem as an unpaired image-to-image translation where the goal is to learn the mapping between an input endoscopic image and a corresponding annotation using an adversarial model. Our approach allows to train image segmentation models without the need to acquire expensive annotations and can potentially exploit large unlabeled endoscopic image collection outside the annotated distributions of image/annotation data. We test
our proposed method on Endovis 2017 challenge dataset and show that it is competitive with supervised segmentation methods.

\end{abstract}

\section{INTRODUCTION}

Robot-assisted Minimally Invasive Surgery (RMIS) provides a surgeon with improved control, facilitating procedures in confined and difficult to access anatomical regions. However, complications due to the reduced field-of-view
provided by the surgical camera limit the surgeon’s ability to self-localize. Computer assisted interventions (CAI) can help a surgeon by integrating additional information. For example, overlaying pre- and intra-operative imaging with the surgical console can provide a surgeon with valuable information which can improve decision making during complex procedures ~\cite{taylor2008medical}. Integrating this data is a complex
task and involves understanding relations between the patient anatomy, operating instruments and surgical camera. Segmentation of the instruments in the camera images is a crucial component of this process and can be used to prevent rendered overlays from occluding the instruments while providing crucial input to instrument tracking frameworks ~\cite{pezzementi2009articulated,allan2014d}.

\begin{figure}
\includegraphics[width=\linewidth]{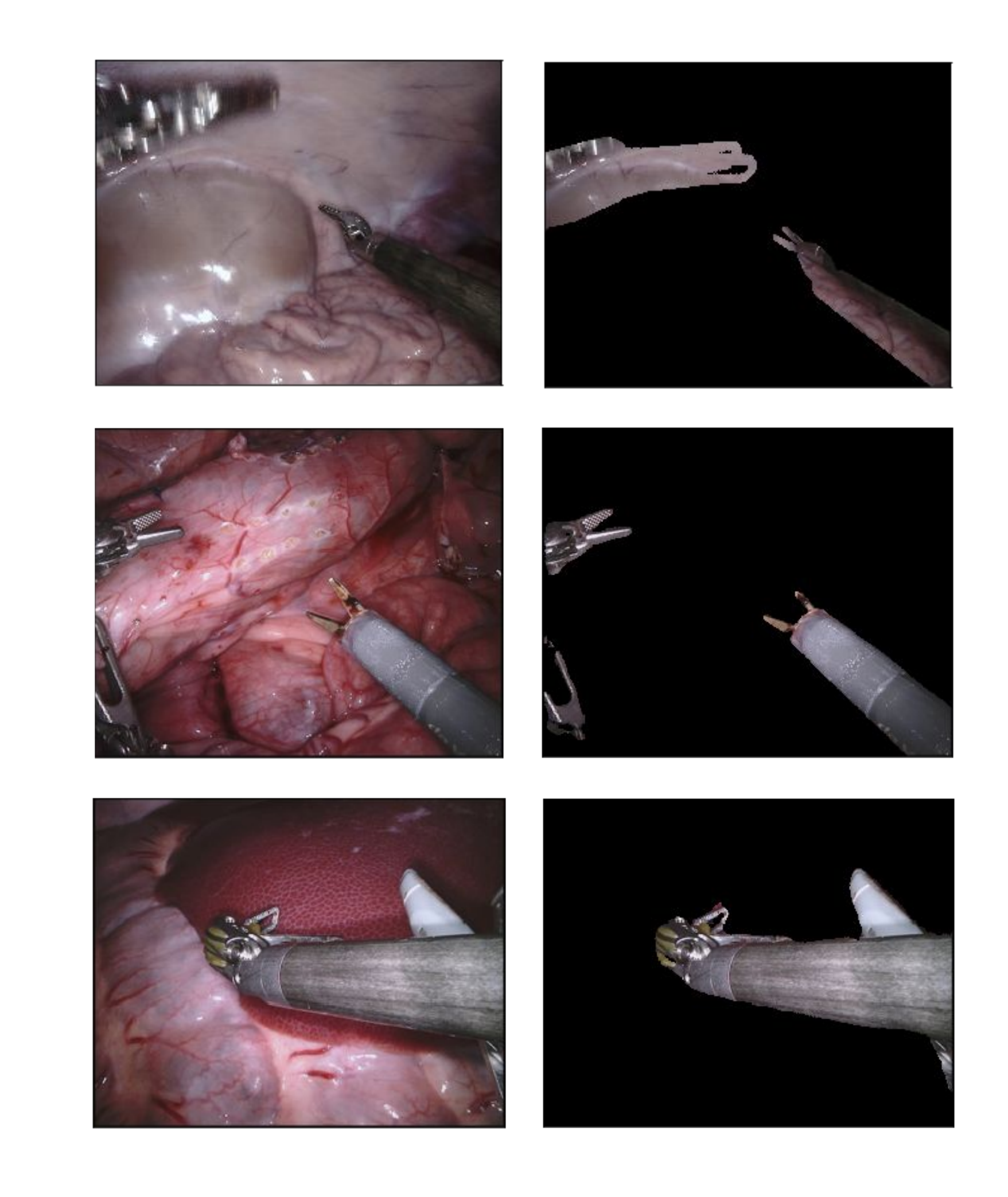}
\caption{\label{fig:demo}Top row shows an example of synthetic annotation acquired using forward kinematics model that was used for training of our method. As it can be seen, it barely captures the actual tool due to errors. Two bottom rows show example segmentations delivered by our method on a random images from Endovis 2017 dataset that were not used for training. }
\label{fig:kinematics_error}
\end{figure}

Segmentation of surgical tools from tissue backgrounds is an extremely difficult task due to lighting challenges such as reflections, shadows and occlusions such as smoke and blood. Early methods attempted to simplify the problem by modifying the appearance of the instruments \cite{tonet2005tracking}. However, this complicates clinical application of the technique as sterilization can become an issue. Segmentation of the instruments using natural appearance is a more desirable approach as it can be applied directly to pre-existing clinical setups. However, this defines a more challenging problem. To solve it, previous work has relied on machine learning techniques to model the complex discriminative boundary. Approaches based on Random Forests \cite{bouget2015detecting}, maximum likelihood Gaussian Mixture Models \cite{pezzementi2009articulated} and Naive Bayesian classifiers \cite{speidel2006tracking}, all trained on color features, have been applied. More recently, the state-of-the-art has increasingly been defined by Fully Convolutional Networks (FCNs), such as the FCN-8s model~\cite{long2015fully} adapted for the task of binary segmentation of robotic tools \cite{garciareal} and U-Net \cite{RonnebergerFB15} which was used for both binary and instrument part segmentation~\cite{laina2017concurrent}. In order for segmentation approaches based on supervised training of neural networks to successfully generalize, a considerable amount of annotated data is required~\cite{cordts2016cityscapes}. At the same time, creating a segmentation dataset with high resolution fine annotations is an extremely time-consuming and costly process~\cite{cordts2016cityscapes}~\cite{allan20192017}.

\begin{figure*}[!h]
  \centering
  \includegraphics[width=\textwidth]{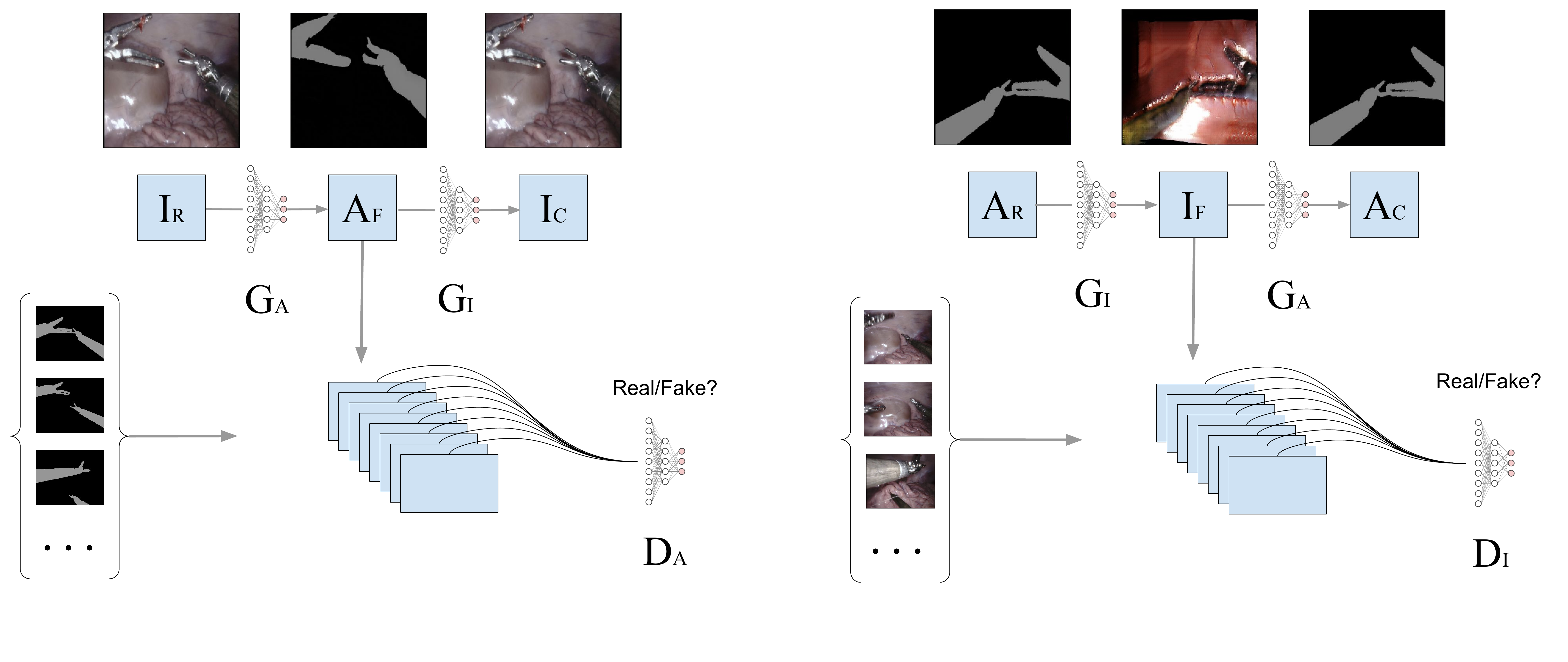}
  \caption{Figure demonstrates main stages of our approach. During training, we learn four mappings which are represented by convolutional neural networks: generators $G_A$ and $G_I$ which map images to annotations and annotations to images respectively; discriminators $D_A$ and $D_I$ which learn to differentiate between real and generated annotations and images respectively. $I_R$ and $A_R$ stand for real images and annotations. $I_F$ and $A_F$ stand for images and annotations generated by networks. $I_C$ and $A_C$ stand for images and annotations generated from $A_F$ and $I_F$ respectively. They are used in cycle-consistency loss term. Loss also includes adversarial terms that forces discriminator to differentiate between real samples and the ones generated by the generators, and the generators to fool the discriminators. After the training, we use $G_A$ to acquire segmentations of unseen images. }

\end{figure*}

Recently, unsupervised and self-supervised methods were introduced to make use of a great amount of unlabeled images of surgical instruments which can be collected with little or no effort: Ross et al.~\cite{ross2018exploiting} proposed a method, which allows to pretrain segmentation models on recoloring task, greatly reducing the number of labeled images necessary for supervised learning; Rocha et al.~\cite{rocha2019self} proposes an optimization method to obtain corrected labels to train a binary segmentation model despite imprecise kinematic model which sometimes results in poor results when the  approximate labels are located far from the actual position of a surgical tool. Apart from instrument segmentation domain, Mahmood et al.~\cite{mahmood2019deep} trained a model for multi-organ nuclei segmentation with synthetic and real data, Pfeiffer et al. ~\cite{pfeiffer2019generating} generated synthetic dataset for liver segmentation and showed promising results.

One way to acquire segmentation masks automatically is to use kinematic model of a robot and CAD model of the tools and project them onto the image plane of camera. Unfortunately, since the kinematic model is imprecise, the generated projections exhibit errors and cannot be directly used for supervised training. Although a lot of effort was made in improving supervised training of instrument segmentation models~\cite{pakhomov2019deep}~\cite{laina2017concurrent}~\cite{garciareal}, a method that is able to successfully use this kind of data for training can potentially be supplied with unlimited amount of automatically generated data.

In this paper, we introduce an approach for binary
surgical tool segmentation which does not need any manual annotation and only uses labels generated with imprecise kinematic model. The problem is posed as image-to-image translation~\cite{zhu2017unpaired} where we need to convert an image of surgical scene from RGB representation to semantic labels representation. Due to errors in the generated annotations that are caused by the imprecise kinematics, we are not able to train a segmentation in a supervised setting using acquired image/annotation pairs. Although we lack direct supervision in the form of image and annotation pairs, we employ set-level supervision: we are given a set of surgical images and a set of generated annotations which were created automatically without any manual annotation (see Fig.~\ref{fig:unpaired}). By using this approach, we learn the mapping from surgical images to binary tool segmentation which is competitive with supervised algorithms that use expensive manually annotated data. This will allow segmentation models to be trained on larger amount of data and generalize better.

\section{METHOD}

\subsection{Data Generation}
 
 In our work we are not using images with  manually created labels similar to Endovis 2017~\cite{allan20192017} for training and, instead, rely only on labels generated with imprecise kinematics. We record image sequences of surgical procedures and corresponding annotation masks. The masks are generated by rendering CAD models of each instrument that is attached to a da Vinci Xi system using joint encoder values that are synchronized to the video feed. Rigid body transforms between the robot base frame and the instruments and the camera are computed using forward kinematics from the DH parameters of the robot. Camera calibration is also acquired from the system. Inaccuracies exist between the true instrument and camera poses due to unknown hand-eye calibration transforms and errors from slack and tension in the cable driven arms of the da Vinci~\cite{allan2014d}.
 
 Overall, three video sequences of different procedures were collected resulting in $6$ thousand frames and annotations. As mentioned previously, annotations exhibit errors similar to example presented in  Fig.~\ref{fig:demo}. The process of the data creation is fully automatic and does not involve any manual annotation and, therefore, can be used to generate a dataset of potentially unlimited size (see Fig.~\ref{fig:unpaired}).
 
Collected video sequences contain 3 different types of robotic surgical instruments: Large Needle Driver, Prograsp Forceps, Bipolar Forceps. An example image sample from the dataset is shown in Fig.~\ref{fig:demo}. Although we were not able to collect video sequences featuring all instruments~\cite{allan20192017} because their CAD models are not available, we noticed that our approach successfully generalizes to previously unseen instruments.
 
\subsection{Set-Level Supervision}
 
 Since the annotations that were automatically generated for our collected images contain errors, we can not use supervision on the level of image/annotation pairs during training. Instead, we use set-level supervision: we propose an unpaired GAN-based approach that learns mappings between surgical images ($I$) and annotation masks ($A$) and enforces them to be cycle consistent. We use cycleGAN~\cite{zhu2017unpaired} approach to learn a mapping $G_A: I \rightarrow A$ between surgical images and corresponding annotation masks that generalizes to previously unseen surgical images.
 
 Overall, our approach employs four networks: $G_I$ (segmentation labels to surgical image generator), $G_A$ (surgical image to segmentation labels generator), $D_A$ (discriminator network of $G_A$), and $D_I$ (discriminator network of $G_I$). The final objective~\cite{zhu2017unpaired} consists of two adversarial loss terms $\mathcal{L}_{\text{GAN}}$ and cycle consistency loss term $\mathcal{L}_{\text{cyc}}$. After the training is done, we use $G_A$ as our segmentation model and evaluate it on previously unseen images to assess its segmentation performance.

First, we introduce an adversarial loss~\cite{zhu2017unpaired} that forces our segmentation model $G_A$ to translate surgical images to realistic segmentation masks by making them look similar to annotations that we collected from kinematics model:

\begin{align}
    \mathcal{L}_{\text{GAN}}(G_A,D_A) =& \ \mathbb{E}_{a \sim p_{\text{data}}(a)}[\log D_A(a)] \nonumber \\
   +& \ \mathbb{E}_{i \sim p_{\text{data}}(i)}[\log (1-D_A(G_A(i))]
\end{align}
 
Where $D_A$ is a discriminator network that learns to distinguish between segmentations generated by our model $G_A$ and real segmentations. Distributions $p_{\text{data}}(i)$ and $p_{\text{data}}(a)$ are approximated by surgical images and annotation samples collected in the prevous step using kinematics model. However, optimizing this objective alone does not guarantee that a meaningful mapping $G_A$ will be learnt since there are infinitely many mappings that will satisfy this criteria~\cite{zhu2017unpaired}.

To make our problem more well-defined we follow~\cite{zhu2017unpaired} and add a constraint on translation to be cycle-consistent: we introduce a reverse mapping $G_I: A \rightarrow I$ from surgical images to corresponding segmentation masks (Fig. 2) and make $G_I$ and $G_A$ to be inverses of each other. Note, that the reverse mapping is introduced only to train our segmentation model $G_A$ more effectively~\cite{zhu2017unpaired}.

 Second adversarial term of loss is introduced to force the distribution of surgical images generated by the network $G_I$ to be similar to that of the real surgical images:
 
 \begin{align}
    \mathcal{L}_{\text{GAN}}(G_I,D_I) =& \ \mathbb{E}_{i \sim p_{\text{data}}(i)}[\log D_I(i)] \nonumber \\
   +& \ \mathbb{E}_{a \sim p_{\text{data}}(a)}[\log (1-D_I(G_I(a))]
\end{align}

 where the mapping $G_I$ is constrained to generate surgical images from real segmentation masks so that they look like real surgical images. Discriminator $D_I$ is trying to differentiate between real and generated surgical images. This term introduces competition between generator and discriminator networks, allowing them to improve during training as a result of the competition~\cite{zhu2017unpaired}.
 
As previously mentioned, in order to make our mapping more well-defined and learn a better segmentation model we also introduce a cycle consistency loss:

\begin{align}
    \mathcal{L}_{\text{cyc}}(G_A, G_I) =  & \ \mathbb{E}_{i\sim p_{\text{data}}(i)}[\norm{G_I(G_A(i))-i}_1] \nonumber \\ 
    + &\ \mathbb{E}_{a\sim p_{\text{data}}(a)}[\norm{G_A(G_I(a))-a}_1]
\end{align}

This term adds an additional regularization by making $G_I$ and $G_A$ inverse functions of each other. Specifically, it ensures that $G_I(G_A(i)) \approx i$ (forward cycle consistency) and $G_A(G_I(a)) \approx a$ (backwards cycle consistency).

The full objective for our set-level supervised learning can be written as:

\begin{align}
     \mathcal{L}(G_A,G_I,D_A,D_I) = & \mathcal{L}_{\text{GAN}}(G_A,D_A) \nonumber \\
    +&\ \mathcal{L}_{\text{GAN}}(G_I,D_I) \nonumber \\
    +& \  \lambda \mathcal{L}_{\text{cyc}}(G_A, G_I)
\end{align}
Where $\lambda$ is a hyperparameter. We aim to solve:
\begin{equation}
    G_A^*,G_I^* = \arg\min_{G_A,G_I}\max_{D_A,D_I} \mathcal{L}(G_A, G_I, D_A, D_I)
\end{equation}

\subsection{Edge Consistency}

Applying the aforementioned method as-is allows us to learn the mapping $G_A$ that acts as a segmentation model and delivers realistically looking segmentation masks but upon inspection, they are completely unaligned with 
instruments present in the input image. Intersection over union accuracy measure is also very low, which means that the segmentation has little or no overlap with tools located in the image. This motivates us to introduce additional constraints so that generated annotations are more aligned with instruments located in the surgical images.

\begin{figure}
\includegraphics[width=\linewidth]{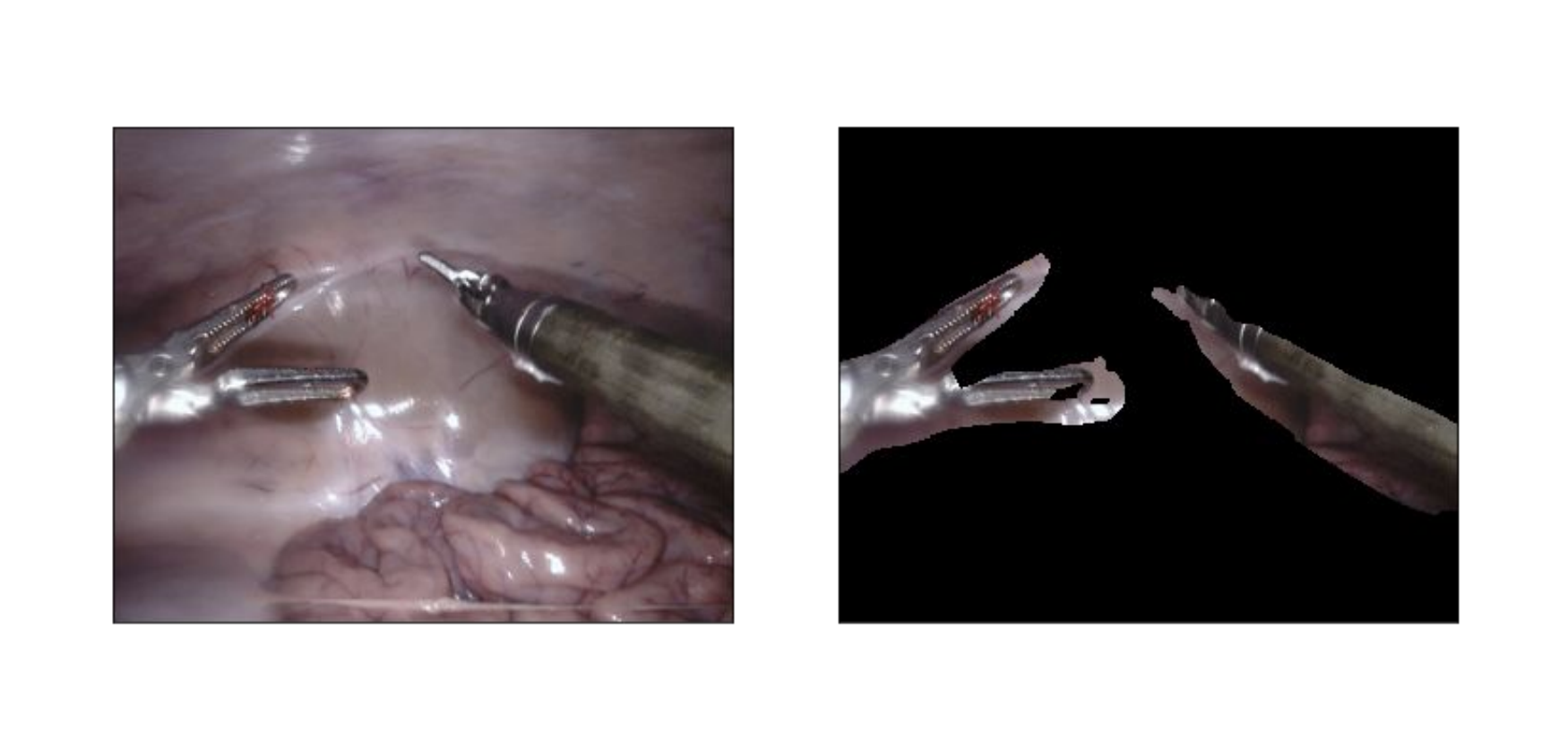}
\caption{\label{fig:poor_results} An example of segmentation results delivered by a deep residual model without edge consistency loss term. As it can be seen, while the shape of the segmentation looks realistic, it is not aligned with the edges of the tools. Since the segmentation result still has a great overlap with the tools, the intersection over union score is good but applications involving augmented reality and tracking of tools require the segmentation algorithm to be more precise along the borders. }
\label{fig:edge_error}
\end{figure}

We replace U-Net~\cite{RonnebergerFB15} architecture that we used for generator networks $G_I$ and $G_A$ with a network based on deep residual connections which give us much better results~\cite{he2016deep}~\cite{zhu2017unpaired}. We hypothesize that an implicit regularization that residual connections provide~\cite{greff2016highway} allows us to acquire segmentation results that are more aligned with the input image. 

However, the problem is not completely solved: while the intersection over union score is good, the predictions delivered by the method are not precise along the borders (See Fig.~\ref{fig:edge_error}). In order for the segmentation method to be useful for augmented reality applications and tool tracking, the segmentation results should be aligned with the borders of the actual tool~\cite{allan20192017}. Ideally, we would want to have consistent edges in the image and generated annotation (See Fig.~\ref{fig:edge_consistency}).
Inspired by a similar problem in the field of image matting~\cite{levinshtein2018real}~\cite{rehemann2009perceptually}, we add the edge consistency term~\cite{levinshtein2018real} to our loss:

\begin{align}
    \mathcal{L}_{\text{edge}}(G_A) = \mathbb{E}_{i \sim p_{\text{data}}(i)}[L_C(G_A(i), i)],
\end{align}

\begin{equation}
L_C(A, I) = \frac{\sum{A_{mag}\big[ 1 - \left( I_x A_x + I_y A_y \right)^2 \big]}}{\sum{A_{mag}}}
\label{eqn:grad_loss}
\end{equation}

where $(I_x, I_y)$ and $(A_x, A_y)$ are the normalized image
and annotation gradients, and $A_{mag}$ is the annotation gradient magnitude. Intuitively, the loss constraints the generator $G_A$ to deliver segmentation masks
that are more aligned with the image edges (See Fig.~\ref{fig:edge_consistency}). Using both the deep residual network and edge consistency gives the best results (See Fig.~\ref{fig:demo}).

\begin{figure}
\includegraphics[width=\linewidth]{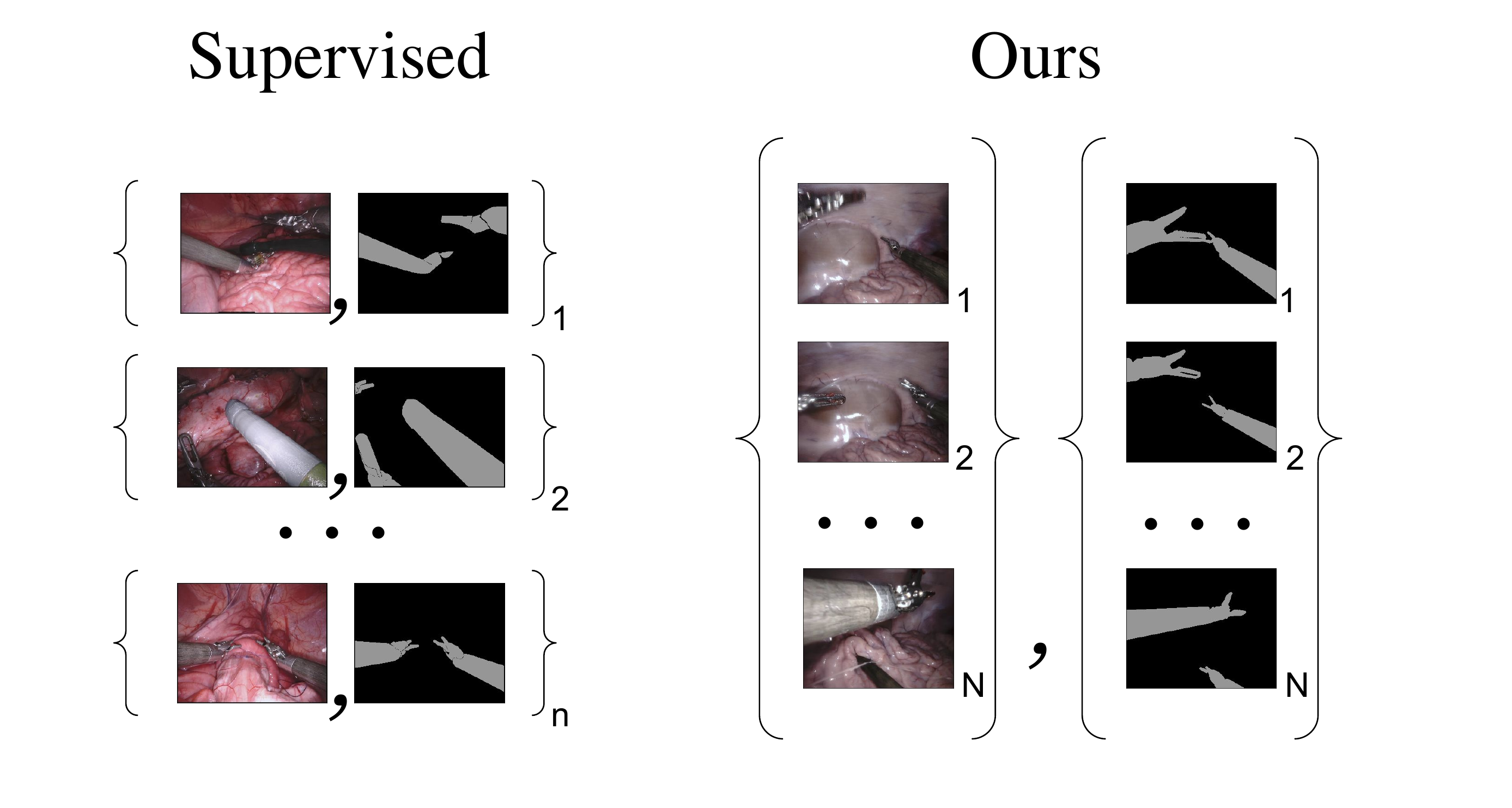}
\caption{\label{fig:unpaired}Supervised learning approach typically uses paired training data (left) consisting of training examples where every image has a corresponding manually annotated image. These datasets are expensive to collect and are usually of a small size. We instead consider unpaired training data (right), consisting of images and synthetically generated annotations using forward kinematics. This data can be automatically generated during medical procedures and can potentially be of unlimited size $N \gg n$.}
\label{fig:kinematics_error}
\end{figure}

\section{EXPERIMENTS AND RESULTS}

\begin{figure}
\includegraphics[width=\linewidth]{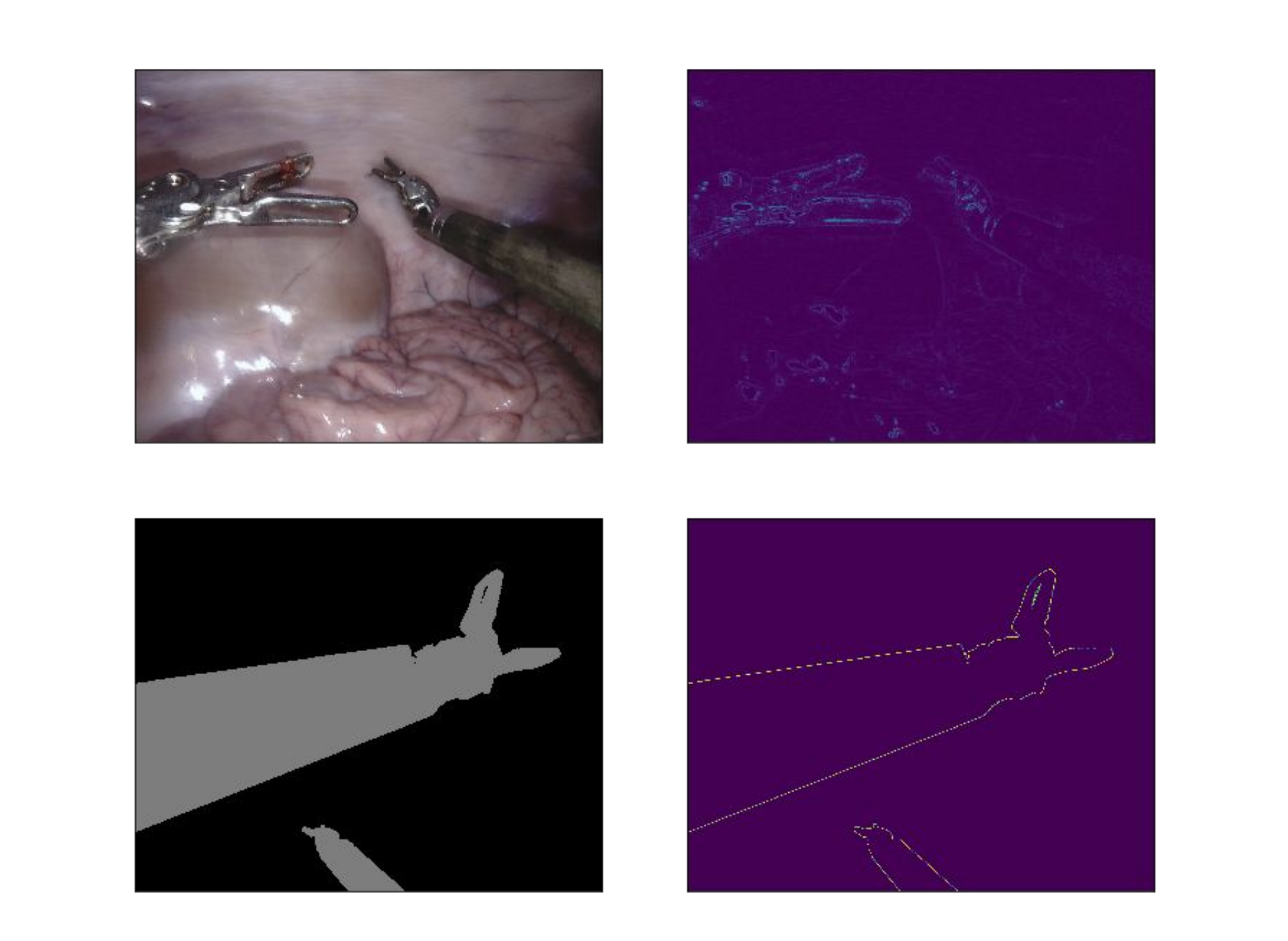}
\caption{\label{fig:edge_example} Examples of surgical image and annotation and their gradient magnitude images. The figures serves as a motivation for edge consistency loss: edges of the image should be aligned with the edges of the segmentation generated by the network. (Image is best viewed in the electronic version of the document) }
\label{fig:edge_consistency}
\end{figure}

\subsection{Implementation Details}
\subsubsection{Network Architectures}

The generator architecture $G_I$ and our segmentation model $G_A$ are both based on deep residual network architecture~\cite{he2016deep}, since it was shown to work better for our task. First two layers of the networks subsample the input image by a factor of $4$, followed by nine residual blocks and two upsampling layers with learnt filters that bring the output to the same size as the input image. The discriminator networks $D_I$ and $D_A$ have much simpler architectures based on PatchGANs~\cite{isola2017image} with three layers and fewer parameters, as suggested in \cite{zhu2017unpaired}. All networks are implemented in a fully-convolutional fashion, which allows them to be applied to images of varying sizes~\cite{long2015fully}.

 In order to reduce memory consumption and be able to store four networks in GPU memory at the same time during training we had to resort to training with batch size one. Since training with small batch size with batch normalization is known to be unstable~\cite{ioffe2015batch}, we are employing instance normalization layers instead~\cite{ulyanov2016instance}.

\subsubsection{Training details}

All the networks in our work were trained from scratch, starting with randomly initilized weights. We perform the optimization with Adam optimizer~\cite{kingma2014adam} with batch size one and learning rate of $0.0002$. Overall, we train for $20$ epochs with fixed learning rate and then linearly decay it to zero for another $20$ epochs.

\subsection{Accuracy measures} \label{Accuracy}

There are three commonly used accuracy measures for assessment of image segmentation models \cite{csurka2013good}: 
\begin{enumerate}
\item Overall Pixel accuracy
\item Per-Class accuracy
\item Jaccard Index (Intersection-over-union)
\end{enumerate}

In order to describe accuracy measures we define confusion matrix $\textbf{C}$, which contains predictions for the whole image segmentation dataset $\textbf{D}$:

$$
\textbf{C}_{ij} = \sum_{I \in \textbf{D} } \left| z \in \textbf{I}~such~that~S^{I}_{gt}(z) = i~and~S^{I}_{ps}(z) = j  \right|
$$

Where $S^{I}_{gt}(z)$ is the ground-truth label of pixel $z$ in image $\textbf{I}$, $S^{I}_{ps}(z)$ is a label predicted by a particular algorithm and $|A|$
is the cardinality of the set $A$. We define $\textbf{G}_i = \sum_{j=1}^{L}C_{ij}$, where $L$ is the number of classes and $\textbf{P}_j = \sum_i \textbf{C}_{ij}$

Overall Pixel accuracy measures number of correctly classified pixels:
$$
OP = \frac{\sum_{i=1}^L\textbf{C}_{ii}}{\sum_{i=1}^L\textbf{G}_{i}}
$$
One significant limitation of this measure
is its bias in the presence of very imbalanced classes~\cite{csurka2013good}.
If a dataset has one class that is more present than others and a segmentation model classifies it correctly, while making mistakes on other smaller classes, the value of the measure will not sufficiently represent that.

Per-Class accuracy does the same measurement as Overall Pixel accuracy but solves its problem with unbalanced classes by scaling results of from each class~\cite{csurka2013good}:
$$
PC =\frac{1}{L}\sum_{i=1}^L  \frac{\textbf{C}_{ii}}{\textbf{G}_{i}}
$$
While it solves the problem, another weakness can be observed: if a large background class is present, one achieves a better score by labeling object classes correctly while making more errors in labeling the background class~\cite{csurka2013good}.

Jaccard Index (Intersection-over-union) measures intersection over union for each class and reports the average between all classes:
$$
JI = \frac{1}{L}\sum_{i=1}^L \frac{\textbf{C}_{ii}}{ \textbf{G }_i+\textbf{P}_i-\textbf{C}_{ii} }
$$
It solves the problem of previous two measures and currently is a main measure of semantic segmentation accuracy for PASCAL VOC challenge~\cite{everingham2010pascal} and Endovis 2017 Robotic Instrument Segmentation Challenge~\cite{allan20192017} which we use to assess our method. In order to be able to compare our method with other methods on Endovis 2017 challenge we report Intersection over Union accuracy measure of our method on test dataset.

\subsection{Dataset}

\begin{table*}[hbt!]
\centering
\small
\scalebox{1.0}{
\begin{tabular}{ l |  c | c | c | c | c | c | c | c | c | c | c | c }
 & NCT & UB & BIT & MIT & SIAT & UCL & TUM & Delhi & UA & UW & Ours & +Edge \\
\hline
Dataset 1  & 0.784  & 0.807  & 0.275  & 0.854  & 0.625  & 0.631  & 0.760  & 0.408  & 0.413  & 0.337 & 0.692 & 0.727 \\
Dataset 2  & 0.788  & 0.806  & 0.282  & 0.794  & 0.669  & 0.645  & 0.799  & 0.524  & 0.463  & 0.289 & 0.735 & 0.769 \\
Dataset 3  & 0.926  & 0.914  & 0.455  & 0.949  & 0.897  & 0.895  & 0.916  & 0.743  & 0.703  & 0.483 & 0.721 & 0.755 \\
Dataset 4  & 0.934  & 0.925  & 0.310  & 0.949  & 0.907  & 0.883  & 0.915  & 0.782  & 0.751  & 0.678 & 0.752 & 0.782 \\
Dataset 5  & 0.701  & 0.740  & 0.220  & 0.862  & 0.604  & 0.719  & 0.810  & 0.528  & 0.375  & 0.219 & 0.778 & 0.794 \\
Dataset 6  & 0.876  & 0.890  & 0.338  & 0.922  & 0.843  & 0.852  & 0.873  & 0.292  & 0.667  & 0.619 & 0.793 & 0.815 \\
Dataset 7  & 0.846  & 0.930  & 0.404  & 0.856  & 0.832  & 0.710  & 0.844  & 0.593  & 0.362  & 0.325 & 0.686 & 0.694 \\
Dataset 8  & 0.881  & 0.904  & 0.366  & 0.937  & 0.513  & 0.517  & 0.895  & 0.562  & 0.797  & 0.506 & 0.787 & 0.815 \\
Dataset 9  & 0.789  & 0.855  & 0.236  & 0.865  & 0.839  & 0.808  & 0.877  & 0.626  & 0.539  & 0.377 & 0.673 & 0.726 \\
Dataset 10  & 0.899  & 0.917  & 0.403  & 0.905  & 0.899  & 0.869  & 0.909  & 0.715  & 0.689  & 0.603 & 0.706 & 0.727 \\
\hline
Mean IOU & 0.843  & 0.875  & 0.326  & 0.888  & 0.803  & 0.785  & 0.873  & 0.612  & 0.591  & 0.461 & 0.732 & 0.760
\end{tabular}
}
\caption{\label{fig:binary_numerical_results} Quantitative results of our method and comparison with supervised methods in binary segmentation of robotic tools~\cite{allan20192017}.}
\end{table*}

We test our method on the EndoVis 2017 Robotic Instruments dataset~\cite{allan20192017}. There are $8$ high resolution ($1280\times1024$) sequences with $225$ frames each in the training dataset~\cite{allan20192017}. As mentioned previously, we did not use the training dataset of Endovis 2017 and, instead, trained our method on our data acquired with imprecise kinematics model.  Each pixel is labeled as either tool or background. 
There are $10$ $75$-frame sequences in the test dataset that features 7 different robotic surgical instruments~\cite{allan20192017}. 
Samples from the dataset and segmentations delivered by our algorithm are depicted in Fig.~\ref{fig:demo}.

The dataset is very challenging and even some of the supervised image segmentation methods were struggling to achieve good performance as it can be seen from the Table.~\ref{fig:binary_numerical_results}. At the same time it provides a good indication of whether our method can generalize well to unseen surgical scenes and instruments. After training our segmentation model on collected images and annotations generated with imprecise kinematics model, we evaluated it on the test set of the Endovis dataset.

\subsection{Quantitative Study and Results}

We used data from ten different test sequences of Endovis 2017 dataset to evaluate our trained instrument segmentation network. To assess generalizability of the developed algorithm, we also payed attention on how our method performed segmentation with instruments that were not represented in our collected dataset: our method successfully segmented previously unseen instrument (See bottom row of Fig.~\ref{fig:demo}). Table~\ref{fig:binary_numerical_results} summarizes the quantitative results of the testing with intersection-over-union metric.

A comparative analysis with  supervised segmentation methods that participated in the challenge was performed and our method, as can be seen from the table, outperforms four out of ten methods. We are the first ones to test a self-supervised method on challenging Endovis 2017 competition while outperforming some of the supervised methods that reported their results.

\section{CONCLUSION AND FUTURE WORK}

Automated training of accurate instrument segmentation models for surgical procedures has the potential to
completely eliminate costs associated with manual creation of datasets and can greatly affect the field by improving the segmentation performance and robustness of segmentation models by employing abundant unlabeled data. In this work we address the problem of training a segmentation model without direct supervision where images and inaccurate labels are generated automatically, therefore, eliminating the need for dataset creation.

We propose an approach that allows the instrument segmentation network to be trained on images with synthetically generated annotations with errors. The problem is posed as an unpaired image-to-image translation task. This way we are able to enforce set-level supervision between sets of surgical images and annotations. This approach performs on par with some standard supervised approaches tested on challenging Endovis 2017 dataset.

In the future work, we plan on adapting this approach to multi-class instrument segmentation and instance segmentation since this data can be easily automatically generated in a similar way to how we generated binary masks. Potentially, other types of mappings can also be learnt without direct supervision that can be very useful for surgical scene analysis and pose estimation like pixel-wise depth estimation and surgical tools landmarks detection. On the other hand, since our approach still performs worse than some of the supervised methods in terms of accuracy, a better segmentation network architecture can be used to close this gap in accuracy and completely eliminate the need for manual segmentation dataset creation.


\begin{thebibliography}{10}

\bibitem{allan2014d}
Max Allan, Ping-Lin Chang, S{\'e}bastien Ourselin, David~J Hawkes, Ashwin
  Sridhar, John Kelly, and Danail Stoyanov.
\newblock Image based surgical instrument pose estimation with multi-class
  labelling and optical flow.
\newblock In {\em International Conference on Medical Image Computing and
  Computer-Assisted Intervention}, pages 331--338. Springer, 2015.

\bibitem{allan20192017}
Max Allan, Alex Shvets, Thomas Kurmann, Zichen Zhang, Rahul Duggal, Yun-Hsuan
  Su, Nicola Rieke, Iro Laina, Niveditha Kalavakonda, Sebastian Bodenstedt,
  et~al.
\newblock 2017 robotic instrument segmentation challenge.
\newblock {\em arXiv preprint arXiv:1902.06426}, 2019.

\bibitem{bouget2015detecting}
David Bouget, Rodrigo Benenson, Mohamed Omran, Laurent Riffaud, Bernt Schiele,
  and Pierre Jannin.
\newblock Detecting surgical tools by modelling local appearance and global
  shape.
\newblock {\em IEEE transactions on medical imaging}, 34(12):2603--2617, 2015.

\bibitem{cordts2016cityscapes}
Marius Cordts, Mohamed Omran, Sebastian Ramos, Timo Rehfeld, Markus Enzweiler,
  Rodrigo Benenson, Uwe Franke, Stefan Roth, and Bernt Schiele.
\newblock The cityscapes dataset for semantic urban scene understanding.
\newblock In {\em Proceedings of the IEEE conference on computer vision and
  pattern recognition}, pages 3213--3223, 2016.

\bibitem{csurka2013good}
Gabriela Csurka, Diane Larlus, Florent Perronnin, and France Meylan.
\newblock What is a good evaluation measure for semantic segmentation?.
\newblock In {\em BMVC}, volume~27, page 2013. Citeseer, 2013.

\bibitem{everingham2010pascal}
Mark Everingham, Luc Van~Gool, Christopher~KI Williams, John Winn, and Andrew
  Zisserman.
\newblock The pascal visual object classes (voc) challenge.
\newblock {\em International journal of computer vision}, 88(2):303--338, 2010.

\bibitem{garciareal}
L.~C. Garc{\i}a-Peraza-Herrera, W.~Li, C.~Gruijthuijsen, A.~Devreker,
  G.~Attilakos, J.~Deprest, E.~Vander~Poorten, D.~Stoyanov, T.~Vercauteren, and
  S.~Ourselin.
\newblock Real-time segmentation of non-rigid surgical tools based on deep
  learning and tracking.
\newblock In {\em CARE Workshop (MICCAI)}, 2016.

\bibitem{greff2016highway}
Klaus Greff, Rupesh~K Srivastava, and J{\"u}rgen Schmidhuber.
\newblock Highway and residual networks learn unrolled iterative estimation.
\newblock {\em arXiv preprint arXiv:1612.07771}, 2016.

\bibitem{he2016deep}
Kaiming He, Xiangyu Zhang, Shaoqing Ren, and Jian Sun.
\newblock Deep residual learning for image recognition.
\newblock In {\em Proceedings of the IEEE Conference on Computer Vision and
  Pattern Recognition}, pages 770--778, 2016.

\bibitem{ioffe2015batch}
Sergey Ioffe and Christian Szegedy.
\newblock Batch normalization: Accelerating deep network training by reducing
  internal covariate shift.
\newblock {\em arXiv preprint arXiv:1502.03167}, 2015.

\bibitem{isola2017image}
Phillip Isola, Jun-Yan Zhu, Tinghui Zhou, and Alexei~A Efros.
\newblock Image-to-image translation with conditional adversarial networks.
\newblock In {\em Proceedings of the IEEE conference on computer vision and
  pattern recognition}, pages 1125--1134, 2017.

\bibitem{kingma2014adam}
Diederik Kingma and Jimmy Ba.
\newblock Adam: A method for stochastic optimization.
\newblock {\em arXiv preprint arXiv:1412.6980}, 2014.

\bibitem{laina2017concurrent}
Iro Laina, Nicola Rieke, Christian Rupprecht, Josu{\'e}~Page Vizca{\'\i}no,
  Abouzar Eslami, Federico Tombari, and Nassir Navab.
\newblock Concurrent segmentation and localization for tracking of surgical
  instruments.
\newblock In {\em International Conference on Medical Image Computing and
  Computer-Assisted Intervention}, pages 664--672, 2017.

\bibitem{levinshtein2018real}
Alex Levinshtein, Cheng Chang, Edmund Phung, Irina Kezele, Wenzhangzhi Guo, and
  Parham Aarabi.
\newblock Real-time deep hair matting on mobile devices.
\newblock In {\em 2018 15th Conference on Computer and Robot Vision (CRV)},
  pages 1--7. IEEE, 2018.

\bibitem{long2015fully}
Jonathan Long, Evan Shelhamer, and Trevor Darrell.
\newblock Fully convolutional networks for semantic segmentation.
\newblock In {\em Proceedings of the IEEE Conference on Computer Vision and
  Pattern Recognition}, pages 3431--3440, 2015.

\bibitem{mahmood2019deep}
Faisal Mahmood, Daniel Borders, Richard Chen, Gregory~N McKay, Kevan~J
  Salimian, Alexander Baras, and Nicholas~J Durr.
\newblock Deep adversarial training for multi-organ nuclei segmentation in
  histopathology images.
\newblock {\em IEEE transactions on medical imaging}, 2019.

\bibitem{pakhomov2019deep}
Daniil Pakhomov, Vittal Premachandran, Max Allan, Mahdi Azizian, and Nassir
  Navab.
\newblock Deep residual learning for instrument segmentation in robotic
  surgery.
\newblock In {\em International Workshop on Machine Learning in Medical
  Imaging}, pages 566--573. Springer, 2019.

\bibitem{pezzementi2009articulated}
Zachary Pezzementi, Sandrine Voros, and Gregory~D Hager.
\newblock Articulated object tracking by rendering consistent appearance parts.
\newblock In {\em Robotics and Automation, 2009. ICRA'09. IEEE International
  Conference on}, pages 3940--3947. IEEE, 2009.

\bibitem{pfeiffer2019generating}
Micha Pfeiffer, Isabel Funke, Maria~R Robu, Sebastian Bodenstedt, Leon
  Strenger, Sandy Engelhardt, Tobias Ro{\ss}, Matthew~J Clarkson, Kurinchi
  Gurusamy, Brian~R Davidson, et~al.
\newblock Generating large labeled data sets for laparoscopic image processing
  tasks using unpaired image-to-image translation.
\newblock In {\em International Conference on Medical Image Computing and
  Computer-Assisted Intervention}, pages 119--127. Springer, 2019.

\bibitem{rehemann2009perceptually}
Christoph Rhemann, Carsten Rother, Jue Wang, Margrit Gelautz, Pushmeet Kohli,
  and Pamela Rott.
\newblock A perceptually motivated online benchmark for image matting.
\newblock In {\em 2009 IEEE Conference on Computer Vision and Pattern
  Recognition}, pages 1826--1833. IEEE, 2009.

\bibitem{rocha2019self}
Cristian da~Costa Rocha, Nicolas Padoy, and Benoit Rosa.
\newblock Self-supervised surgical tool segmentation using kinematic
  information.
\newblock {\em arXiv preprint arXiv:1902.04810}, 2019.

\bibitem{RonnebergerFB15}
Olaf Ronneberger, Philipp Fischer, and Thomas Brox.
\newblock U-net: Convolutional networks for biomedical image segmentation.
\newblock {\em CoRR}, abs/1505.04597, 2015.

\bibitem{ross2018exploiting}
Tobias Ross, David Zimmerer, Anant Vemuri, Fabian Isensee, Manuel Wiesenfarth,
  Sebastian Bodenstedt, Fabian Both, Philip Kessler, Martin Wagner, Beat
  M{\"u}ller, et~al.
\newblock Exploiting the potential of unlabeled endoscopic video data with
  self-supervised learning.
\newblock {\em International journal of computer assisted radiology and
  surgery}, 13(6):925--933, 2018.

\bibitem{speidel2006tracking}
Stefanie Speidel, Michael Delles, Carsten Gutt, and R{\"u}diger Dillmann.
\newblock Tracking of instruments in minimally invasive surgery for surgical
  skill analysis.
\newblock In {\em International Workshop on Medical Imaging and Virtual
  Reality}, pages 148--155. Springer, 2006.

\bibitem{taylor2008medical}
Russell~H Taylor, Arianna Menciassi, Gabor Fichtinger, and Paolo Dario.
\newblock Medical robotics and computer-integrated surgery.
\newblock In {\em Springer handbook of robotics}, pages 1199--1222. Springer,
  2008.

\bibitem{tonet2005tracking}
Oliver Tonet, TU~Ramesh, Giuseppe Megali, and Paolo Dario.
\newblock Tracking endoscopic instruments without localizer: image
  analysis-based approach.
\newblock {\em Studies in health technology and informatics}, 119:544--549,
  2005.

\bibitem{ulyanov2016instance}
Dmitry Ulyanov, Andrea Vedaldi, and Victor Lempitsky.
\newblock Instance normalization: The missing ingredient for fast stylization.
\newblock {\em arXiv preprint arXiv:1607.08022}, 2016.

\bibitem{zhu2017unpaired}
Jun-Yan Zhu, Taesung Park, Phillip Isola, and Alexei~A Efros.
\newblock Unpaired image-to-image translation using cycle-consistent
  adversarial networks.
\newblock In {\em Proceedings of the IEEE international conference on computer
  vision}, pages 2223--2232, 2017.

\end{thebibliography}
\end{document}